\documentclass[twoside,11pt]{article}

\usepackage{jmlr2e}
\usepackage{geometry}
\usepackage[flushleft]{threeparttable}
\usepackage{algpseudocode,algorithm,algorithmicx}
\usepackage{amsmath,amssymb}
\usepackage{soul}
\usepackage{array, makecell}
\usepackage{soul}
\usepackage{color}
\usepackage{longtable}
\usepackage{booktabs}
\usepackage[english]{}
\usepackage{multirow}
\usepackage{subfigure}
\usepackage{multicol}
\usepackage{xcolor}
\usepackage{ragged2e}
\usepackage{changepage}
\usepackage{graphicx} 
\usepackage{rotating}

\usepackage{amsfonts,amssymb,amsbsy,tabulary,amsmath}
\usepackage{mathtools}
\usepackage{soul}
\usepackage{graphicx}
\usepackage{lineno}
\usepackage{prodint}

\usepackage[normalem]{ulem} 
\usepackage{algorithm}
\usepackage{amsmath}
\usepackage{caption}
\usepackage{listings}
\usepackage{color} 

\usepackage[T1]{fontenc}
\usepackage{bigfoot} 
\usepackage[]{matlab-prettifier}
\usepackage{multicol}
\usepackage{multirow}
\usepackage{filecontents}

\hypersetup{
    citecolor=red,
    linkcolor=red,   
    urlcolor=Magenta}
\usepackage{tcolorbox}

\lstMakeShortInline"

\firstpageno{1}

\begin{document}
\title{ A Metric-based Principal Curve Approach for Learning One-dimensional Manifold}

\author{\name Eliuvish Cuicizion \email                           elviscuihan@g.ucla.edu\\
       \addr Department of Biostatistics\\
       University of California, Los Angeles, CA 90095, USA \\
       }

\maketitle

\begin{abstract}
Principal curve is a well-known statistical method oriented in manifold learning using concepts from differential geometry. In this paper, we propose a novel metric-based principal curve (MPC) method that learns one-dimensional manifold of spatial data. Synthetic datasets  Real applications using MNIST dataset show that our method can learn the one-dimensional manifold well in terms of the shape.

\end{abstract}
\medskip


\section{A Brief Review on Differential Geometry in Statistics}

The application of differential geometry in statistics should be credit to two great Indian statisticians \cite{mahalanobis2018generalized} (the original paper was published in 1936) and \cite{rao1945information}. One of the early pioneer papers of differential geometry in statistics was done by \cite{efron1975defining}. He defined the concept \emph{statistical curvature} rigorously for the first time. Following his work, \cite{skovgaard1984riemannian} studied the Riemannian geometry of a family of multivariate normal models in depth. Some other early works include \cite{efron1978geometry, atkinson1981rao, amari1982differential, campbell1985relation, barndorff1986role} and \cite{ravishanker1990differential}. Two excellent monographs in connecting differential geometry and statistics are \cite{murray1993differential} and \cite{amari1987differential}. In addition, the concept of \textit{tangent space} is borrowed from differential geometry and deeply studied in \cite{bickel1993efficient}.
Apart from the conventional work mentioned above, there are also much more interesting and exciting progress going on in bridging these two fields (which are under the fancy names \textit{metric learning} \citep{izenman2008modern}, \textit{topological data analysis} \citep{rabadan2019topological}, \textit{directional statistics} \citep{mardia2000directional}, \textit{shape analysis} \citep{bhattacharya2012nonparametric, dryden2016statistical} and \textit{functional data analysis} \citep{wang2016functional} , etc.). In most of these works, data are not assumed to be sampled from a regular Euclidean space (say, $\mathbb{R}^d$) but a smooth low-dimensional manifold embedded in $\mathbb{R}^d$ \citep{do1992riemannian, tapp2016differential}. To the best of our knowledge, it was \cite{hastie1984principal} who first defined such a manifold in a statistical setting and he named it as the \textit{principal curve}. His idea was later developed in \cite{hastie1989principal} and \cite{tibshirani1992principal} and more recently in \cite{ozertem2011locally}. We list some of other applications of differential geometry methods in statistics in table~\ref{tab:differential_geometry}. For a more comprehensive review, we suggest the article written by \cite{wasserman2018topological}.

\begin{longtable}{p{0.55\linewidth}r}
    \caption{\large Applications of Differential Geometry in Statistics}\\
\toprule
\textbf{Method / Approach} \textsuperscript{1} & \textbf{Reference} \\ 
\midrule 
Principle curves and surfaces & \cite{hastie1989principal}\\
Kernel principal component analysis (K-PCA) & \cite{scholkopf1997kernel} \\
Local linear embedding (LLE) & \cite{roweis2000nonlinear}\\
Isometric feature mapping (ISOMAP) & \cite{tenenbaum2000global} \\
Laplacian eigenmap & \cite{belkin2003laplacian} \\
Hessian eigenmaps & \cite{donoho2003hessian} \\ 
Intrinsic dimension estimation (IDE) & \cite{levina2004maximum} \\
Principal geodesic analysis (PGA)&\cite{fletcher2004principal} \\
Intrinsic statistics on Riemannian manifolds & \cite{pennec2006intrinsic}\\
Nonparametric regression on Riemannian manifolds & \cite{pelletier2006non} \\
The diffusion maps & \cite{coifman2006diffusion}\\
Shape-space smoothing & \cite{kume2007shape} \\
Mapper algorithm & \cite{singh2007topological} \\
Riemannian K-means & \cite{goh2008clustering, zhang2020k}\\
Locally defined principal curves &\cite{ozertem2011locally}\\
Computation of Vietoris-Rips filtration & \cite{sheehy2012linear}\\
Probablistic principal geodesic analysis (P-PGA) & \cite{zhang2013probabilistic} \\
Geodesic mixture models (GMM) & \cite{simo2014geodesic}\\
Geodesic convolutional neural network (G-CNN) & \cite{masci2015geodesic}\\
Dirichlet process mixture on spherical manifold & \cite{straub2015dirichlet}\\
Locally adaptive normal distribution (LAND) & \cite{arvanitidis2016locally} \\
Uniform manifold approximation and projection & \cite{mcinnes2018umap}\\
Statistical inference on Lie groups & \cite{falorsi2019reparameterizing}\\
Fréchet regression & \cite{petersen2019frechet}\\
Geometrically enriched latent space approach &\cite{arvanitidis2020geometrically}\\
Wasserstein regression & \cite{chen2021wasserstein, matabuena2021glucodensities} \\
Pseudotime analysis & \cite{cui2022single}\\
Invertible kernel PCA (IK-PCA) & \cite{gedon2023invertible} \\
\bottomrule
\label{tab:differential_geometry}
\end{longtable}
\vspace{-5mm}
\begin{minipage}{\linewidth}
\textsuperscript{1} \footnotesize This table only includes methods that use concepts or tools from differential geometry. Hence, some other popular manifold learning techniques are not included, such as local discriminant analysis \citep{hastie1995discriminant}, random projection \citep{johnson1984extensions} and t-SNE \citep{van2008visualizing}.
\end{minipage}
\textcolor{white}{.}\newline
We note that many methods listed above are published in non-statistical journals. Hence, it is urgent (and also a necessity) for statisticians to develop concepts, tools and methods that can adapt today's societal needs. Further, the \textit{Geomstats} package \citep{miolane2020geomstats, miolane2020introduction}, the \textit{Geoopt} package \citep{kochurov2020geoopt} and the \textit{Pymanopt} package  \citep{townsend2016pymanopt} in Python and the \textit{umap} package, the \textit{geomorph} package \citep{adams2013geomorph} and the \textit{frechet} package in R \citep{konopka2018r} provides many useful computational tools for the differential geometry methods mentioned in the table. 

The rest of the paper is organized as follows. In section~\ref{sec:metric-pc}, we propose a new approach for learning one-dimensional representation of data and term it as the metric-based principal curve. Simulation studies using synthetic datasets are presented in section~\ref{sec:simu} and real applications using the MNIST dataset is given in section~\ref{sec:mnist}. In appendix~\ref{sec:prelim}, we provide some preliminaries in Riemann geometry.

\section{Metric-based Principal Curve}\label{sec:metric-pc}

In this section, we propose a new algorithm for learning a one-dimensional representation of data. We term it the \textit{metric-based principal curve} as it minimizes a metric distance between the raw data and the projected data using smoothing and regression techniques. We give a formal and rigorous introduction below.

\begin{definition}[Principal curve assumption]
    Let $\mathbf{Y}=(Y_1,Y_2,\cdots,Y_p)^T\in\mathbb{R}^p$ be a $p$-dimensional random vector and $\lambda\in\mathbb{R}$ be a scalar known as the \textbf{projection index}. Extending the idea in \cite{tibshirani1992principal}, the \textbf{principal curve assumption} for $(\mathbf{Y},\lambda)$ is described as follows:
\begin{align}
Y_j|\lambda&\sim_{ind}\mathbb{P}_{Y_j|\lambda},j=1,\cdots,p,\\
\mathcal{P}(\lambda)&=\mathbb{E}\left(\mathbf{Y}|\lambda\right)=\int_{\mathbb{R}^p} y\ \mathbb{P}_{\mathbf{Y}|\lambda}(dy)
\end{align}
where  $\mathbb{P}$ represents a general probability measure and the expectation $\mathcal{P}(\lambda)$ is defined as the \textbf{principal curve} of $\mathbf{Y}$. Alternatively, using the idea of nonparametric regression, we can also assume that
$$Y_j=f_j(\lambda)+\epsilon_j,\ j=1,2,\cdots,p$$
where $f_j(\cdot)$ is a smooth function and $\epsilon_j$ is a random error.
\end{definition}

Note that the above definition is in the population level, i.e., we have the oracle knowledge $\mathbb{P}_{\mathbf{Y}|\lambda}$. However, in practice, we do not have the access to it and have to estimate either $\mathbb{P}_{\mathbf{Y}|\lambda}$ or $f_i$ from data as well as the projection index $\lambda$. There are several projection-expectation (PE) type algorithms to learn the principal curve $\mathcal{P}(\lambda)$ \citep{hastie1989principal, chang1998principal}. However, it is well-known that PE algorithms do not guarantee convergence and different types of definition and algorithm lead to different estimated principal curve \citep{gerber2013regularization}. Hence, it is of great interest to develop new algorithms to estimate principal curves. In the following, we propose a metric-based algorithm for learning principal curves and term the estimator \textbf{the metric-based principal curve} (MPC).

The idea of MPC starts with a user-specified metric $d(\cdot,\cdot)$ on $\mathbb{R}\times\mathbb{R}$ and a regularization parameter $\rho$. Suppose we observe the data $\mathbf{Y}_1,\mathbf{Y}_2,\cdots,\mathbf{Y}_n$ where each $\mathbf{Y}_i\in\mathbb{R}^p$. The associated projection index $\lambda_1,\cdots,\lambda_n$ is chosen such that the following quantity is minimized.
\begin{align}
    \{\lambda_i\}_{i=1}^n&=\arg\min_{\lambda}\frac{1}{n}\sum_{i=1}^nd(\mathbf{Y}_i,\mathbf{\widehat{Y}}(\lambda_i)) + \rho\ \phi(\{\lambda_i\}_{i=1}^n)
\end{align}
 where $\mathbf{\widehat{Y}}(\lambda_i)=(\widehat{f}_{1}(\lambda_i),\widehat{f}_{2}(\lambda_i),\cdots,\widehat{f}_{p}(\lambda_i))^T$ is the fitted value of $\mathbf{Y}_i$ using \textit{under-smooth regression models} $\{f_{j}\}_{j=1}^p$ and $\phi$ is a dispersion function characterizing the dispersion of $\lambda$'s. We provide a list of the smoother $f_j$, the metric $d(\cdot,\cdot),$ and the dispersion $\phi$ in table~\ref{tab:choices}.

 \begin{table}[!ht]
  \caption{Some choices of $f_j$, $d(\cdot,\cdot)$ and $\phi(\cdot)$.}
  \centering 
  \begin{threeparttable}
    \begin{tabular}{lll}
    $f_j$  & $d(x,y)$ & $\phi$ \tnote{*}\\
     \midrule\midrule
Smoothing spline  & $L_d$-distance $\lVert x-y\lVert_d$  & $\sum_{i=1}^{n-1}|\lambda_{(i+1)}-\lambda_{(i)}|$                       \\
    \cmidrule(l  r ){1-3}
     LOWESS & Mahalanobis distance & $\sum_{i=1}^{n-1}\left(\lambda_{(i+1)}-\lambda_{(i)}\right)^2$ \\ 
    \cmidrule(l r ){1-3}
     Kernel ridge regression &  Chebyshev distance & $\max_i|\lambda_{(i+1)}-\lambda_{(i)}|$ \\ 
    \cmidrule(l r ){1-3}
    Gaussian process regression
    &  Hellinger distance   & Coefficient of variation                       \\
    \cmidrule(l  r ){1-3}
    Support vector regression & Canberra distance\\
    \cmidrule(l  r ){1-3}
    Nadaraya-Watson estimator
    \\
    \midrule\midrule
    \end{tabular}
    \begin{tablenotes}
\item[*] Here $\lambda_{(i)}$ denotes the $i^{th}$ order statistics from $\lambda_1,\cdots,\lambda_n$.
\end{tablenotes}
\end{threeparttable}\label{tab:choices}
  \end{table}

In short, a metric-based principal curve minimizes the mean of distances of all points (feature vectors) projected onto the curve plus a regularization term that penalizes the dispersion of the on-dimensional parameter. 


\section{Simulation Studies}\label{sec:simu}
In this section, we perform simulation studies based on three synthetic datasets, namely, spiral curve, golden bridge and Arabic numerals seven.

\subsection{Number Seven}
\begin{figure}[!htb]
\minipage{0.32\textwidth}
  \includegraphics[width=\linewidth]{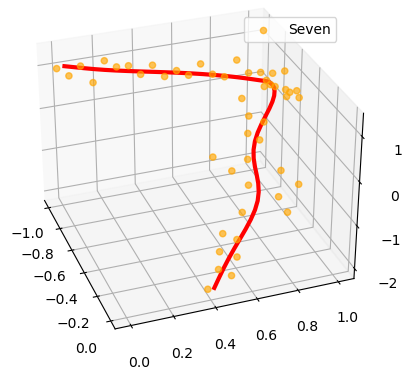}
\endminipage\hfill
\minipage{0.32\textwidth}
  \includegraphics[width=\linewidth]{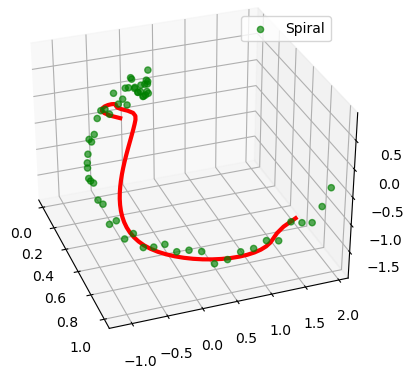}
\endminipage\hfill
\minipage{0.32\textwidth}%
  \includegraphics[width=\linewidth]{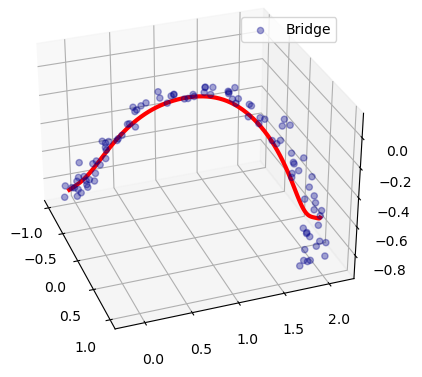}
\endminipage
	\caption{Principal curves of seven, spiral and bridge in $\mathbb{R}^3$. Red lines are learned principal curves which represent the trajectory of data manifold.}
	\label{fig:synthetic}
\end{figure}

The generative model is
\begin{align*}
    Y_1 &=t+\epsilon_1\\
    Y_2 &=U_2^{X}(1+\epsilon_2)^{1-X}\\
    Y_3 &=(1+\epsilon_3)^{X}U_3^{1-X}
\end{align*}
where $t\in[0, 1]$, $U_2\sim\mathcal{U}(0, 1)$, $U_3\sim\mathcal{U}(-2, 0.7)$, $\epsilon_i\sim_{iid}\mathcal{N}(0,0.1)$ and $X\sim \text{Ber}(0.5)$. For simulation, we take the sample size to be 120 and generate $t$ uniformly spaced within $[0, 1]$. For estimation of $\lambda$'s, we set $f_j$ to be smoothing splines, $d(x,y)$ to be $L_2$-distance and $\phi(\lambda)$ to be $\sum_{i=1}^{n-1}|\lambda_{(i+1)}-\lambda_{(i)}|$. For prediction, we set $f_j$ to be LOWESS with bandwidth $0.4$. The results are shown in the left panel of figure~\ref{fig:synthetic}. In addition, we also fit another MPC for $(Y_2,Y_3)$ only since $Y_1$ is just a linear function in $t$. The results are shown in the left panel of figure~\ref{fig:synthetic2}. 

\subsection{Spiral Curve}
The generative model is
\begin{align*}
    Y_1 &=t\\
    Y_2 &=2t\cos\left(6t\right)+\epsilon_2\\
    Y_3 &=2t\sin(6t)+\epsilon_3
\end{align*}
where $\epsilon_i\sim_{iid}\mathcal{N}(0,0.1)$ and $t\in[0, 1]$. For simulation, we take the sample size to be 120 and generate $t$ uniformly spaced within $[0, 1]$. For estimation of $\lambda$'s, we set $f_j$ to be LOWESS with bandwidth 5, $d(x,y)$ to be $L_2$-distance and $\phi(\lambda)$ to be $\sum_{i=1}^{n-1}|\lambda_{(i+1)}-\lambda_{(i)}|$. For prediction, we set $f_j$ to be LOWESS with bandwidth $0.4$. The results are shown in the middle panel of figure~\ref{fig:synthetic}. In addition, we also fit another MPC for $(Y_2,Y_3)$ only since $Y_1$ is just a linear function in $t$. The results are shown in the middle panel of figure~\ref{fig:synthetic2}. 

\subsection{Golden Bridge}

\begin{figure}[!htb]
\minipage{0.32\textwidth}
  \includegraphics[width=\linewidth]{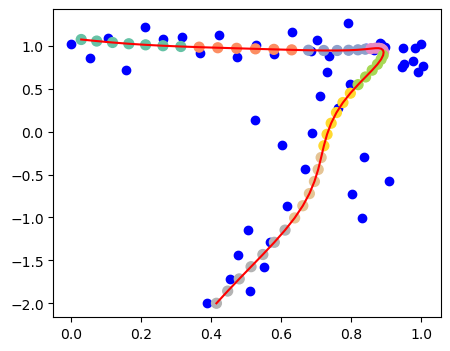}
\endminipage\hfill
\minipage{0.32\textwidth}
  \includegraphics[width=\linewidth]{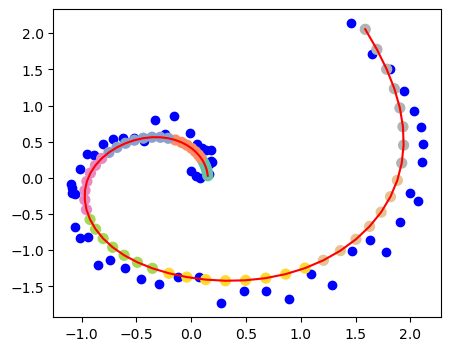}
\endminipage\hfill
\minipage{0.32\textwidth}%
  \includegraphics[width=\linewidth]{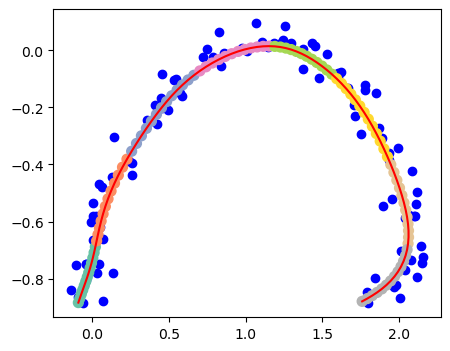}
\endminipage
	\caption{Principal curves of seven, spiral and bridge in $\mathbb{R}^2$. Red lines and colorful points are learned principal curves which represent the trajectory of data manifold. Blue points are raw data in $\mathbb{R}^2$.}
	\label{fig:synthetic2}
\end{figure}

The generative model is
\begin{align*}
    Y_1 &=t\\
    Y_2 &=\sin(2t)+\cos\left(\frac{2}{3}t\right)+\epsilon_2\\
    Y_3 &=-t\sin(2t)+\epsilon_3
\end{align*}
where $\epsilon_i\sim_{iid}\mathcal{N}(0,0.1)$ and $t\in[0, 1]$. For simulation, we take the sample size to be 120 and generate $t$ uniformly spaced within $[0, 1]$. For estimation of $\lambda$'s, we set $f_j$ to be kernel regression with regularization parameter $\alpha=10$, $d(x,y)$ to be $L_2$-distance and $\phi(\lambda)$ to be $\sum_{i=1}^{n-1}|\lambda_{(i+1)}-\lambda_{(i)}|$. For prediction, we set $f_j$ to be smoothing splines. The results are shown in the right panel of figure~\ref{fig:synthetic}. In addition, we also fit another MPC for $(Y_2,Y_3)$ only since $Y_1$ is just a linear function in $t$. The results are shown in the right panel of figure~\ref{fig:synthetic2}. 

\section{Applications to MNIST data}\label{sec:mnist}
\begin{figure}[!htbp]
	\centering
\minipage{0.32\textwidth}
  \includegraphics[width=\linewidth]{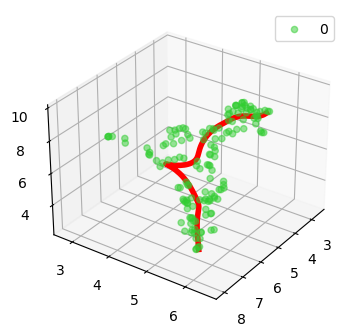}
\endminipage\hfill
\minipage{0.32\textwidth}
  \includegraphics[width=\linewidth]{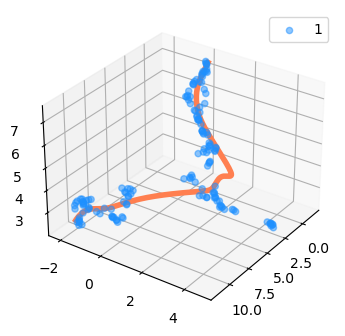}
\endminipage\hfill
\minipage{0.32\textwidth}%
  \includegraphics[width=\linewidth]{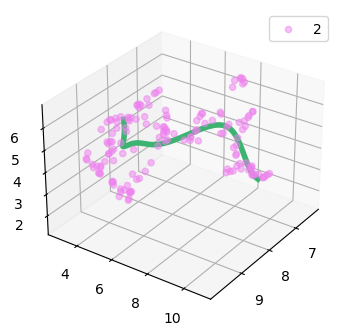}
\endminipage\\
\minipage{0.32\textwidth}
  \includegraphics[width=\linewidth]{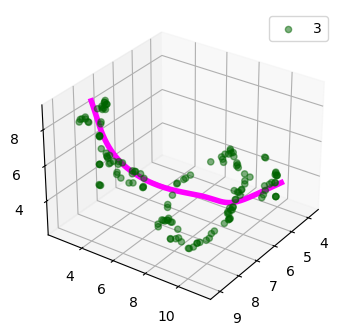}
\endminipage\hfill
\minipage{0.32\textwidth}
  \includegraphics[width=\linewidth]{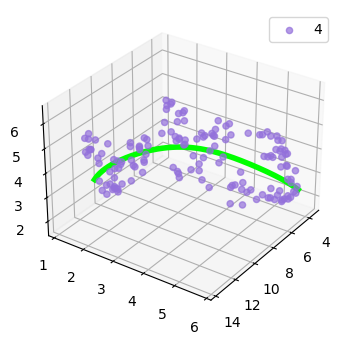}
\endminipage\hfill
\minipage{0.32\textwidth}%
  \includegraphics[width=\linewidth]{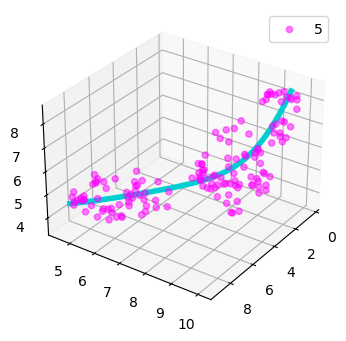}
\endminipage\\
\minipage{0.32\textwidth}
  \includegraphics[width=\linewidth]{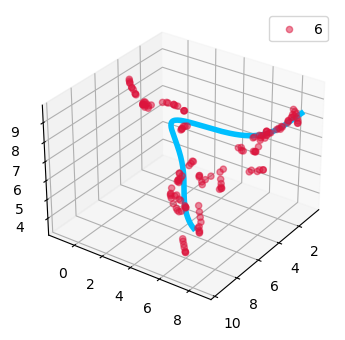}
\endminipage\hfill
\minipage{0.32\textwidth}
  \includegraphics[width=\linewidth]{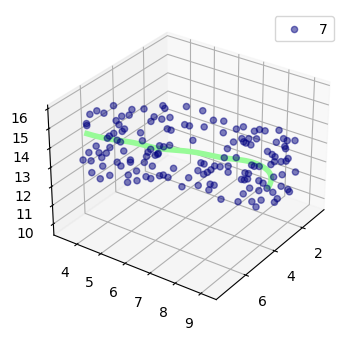}
\endminipage\hfill
\minipage{0.32\textwidth}%
  \includegraphics[width=\linewidth]{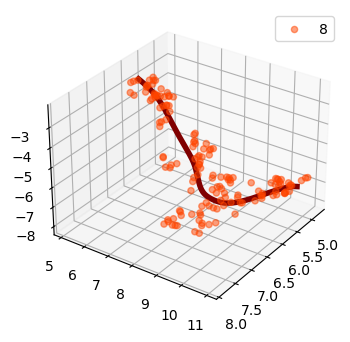}
\endminipage\\
\minipage{0.32\textwidth}%
  \includegraphics[width=\linewidth]{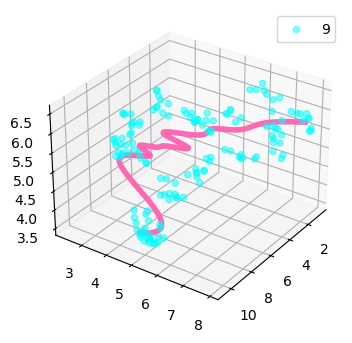}
\endminipage
	\caption{Principal curves of MNIST. Blue lines are learned principal curves which represent the trajectory of data manifold.}
	\label{fig:pc}
\end{figure}

In this section, we apply MPC to the famous MNIST dataset \citep{lecun1998mnist}. We first sample 150 figures for each handwritten digit from the training set. Each figure can be viewed as a point living in a $28\times 28=784$ dimensional space. Next, we apply uniform manifold approximation and projection (UMAP) algorithm to each digit (150 figures) so that each figure is projected onto $\mathbb{R}^3$. Then we perform an MPC analysis to the projected 3-dimensional data for each digit. We visualize the principal curves of all ten digits from in Figure~\ref{fig:pc} for illustration. 

\section{Conclusion}
In this paper, we provide a new approach called Metric-based Principal Curve (MPC) for learning a one-dimensional representation of spatial data points. Such approach constructs a loss function dimension-wise and the minimizer induces an intrinsic order among all points. Theoretical analysis, simulation studies and applications show that our method produces excellent results under different scenarios.

\section{Appendix: Some Preliminaries in Riemann Geometry}\label{sec:prelim}
In this section, we provide some preliminaries on differential geometry for more details, we refer to \cite{do1992riemannian} and \cite{tapp2016differential}. Suppose all points $x\in\mathbb{R}^d$ are column vectors and $\mathcal{S}_{++}^d$ represents the set of symmetric $d\times d$ positive definite matrices. The manifold is denoted $\mathcal{M}$, and its tagent space at $x\in\mathcal{M}$ is denoted $\mathcal{T}_x\mathcal{M}$.
\begin{definition}[Riemannian manifold] A \emph{Riemannian manifold} is a smooth manifold $\mathcal{M}$, equipped with a positive definite \emph{Riemannian metric} $M(x)\ \forall x\in\mathcal{M}$, which is a smoothly varying inner product $\langle v, u\rangle_x=v^TM(x)u$ in the tangent space $\mathcal{T}_x\mathcal{M}$.
\end{definition}
\begin{definition}[Geodesic] Let $\mathcal{M}$ be a Riemannian manifold. A \emph{geodesic} curve $\gamma:[0,1]\rightarrow\mathcal{M}$ is a length-minimizing smmooth curve connecting two given points $x,y\in\mathcal{M}$, i.e.,
\begin{align}
    \gamma(t)&=\arg\min_c L(t,c,c')\\
L(t,c,c')&=\int_0^1\sqrt{c'(t)^TM\left(c(t)\right)c'(t)}dt\\
\gamma(0)&=x\text{ and }\gamma(1)=y\label{eq:boundary_values},
\end{align}
where  $L$ is a functional of $t, c$ and $c'$, $c'(t)\in\mathcal{T}_{c(t)}\mathcal{M}$ is the velocity of the curve $c$ at $t$ and $M$ is the Riemannian metric tensor. Formula~\ref{eq:boundary_values} is referred as \textit{boundary value conditions}.
\end{definition}
\begin{theorem}[Euler-Lagrange equation for geodesic \citep{hauberg2012geometric}] At minima of $L(t,c,c')$, the Euler-Lagrange equation must hold, i.e.,
$$\frac{\partial L}{\partial\gamma}=\frac{d}{dt}\frac{\partial L}{\partial \gamma'}.$$
Hence, geodesic curves embedded in $\mathbb{R}^d$ satisfy the following system of second-order ordinal differential equations (ODE):
\begin{align}
    M(\gamma(t))\gamma''(t)=-\frac{1}{2}\left(\frac{\partial \text{vec}[M(\gamma(t))]}{\partial\gamma(t)}\right)^T\left( \gamma'(t)\otimes\gamma'(t) \right),
\end{align}
where $\otimes$ denotes the Kronecker product and $\text{vec}(\cdot)$ stacks the columns of a matrix into a vector.
    
\end{theorem}

\newpage
\bibliography{sample}

\end{document}